\documentclass[sigconf]{acmart}
\AtBeginDocument{%
  \providecommand\BibTeX{{%
    \normalfont B\kern-0.5em{\scshape i\kern-0.25em b}\kern-0.8em\TeX}}}

\copyrightyear{2024}
\acmYear{2024}
\setcopyright{acmlicensed}\acmConference[FDG 2024]{Proceedings of the 19th International Conference on the Foundations of Digital Games}{May 21--24, 2024}{Worcester, MA, USA}
\acmBooktitle{Proceedings of the 19th International Conference on the Foundations of Digital Games (FDG 2024), May 21--24, 2024, Worcester, MA, USA}
\acmDOI{10.1145/3649921.3659851}
\acmISBN{979-8-4007-0955-5/24/05}

\usepackage{xspace}
\usepackage{caption}
\usepackage{subcaption}
\usepackage{spverbatim}

\usepackage[clock]{ifsym}

\acmConference[FDG '24]{}{May 21--24,
  2024}{Worcester, Massachusetts, USA}
\newcommand*{\mathabxbfamily}{\fontencoding{U}\fontfamily{mathb}\selectfont}
\DeclareFontFamily{U}{mathb}{\hyphenchar\font45}
\DeclareFontShape{U}{mathb}{m}{n}{
      <5> <6> <7> <8> <9> <10> gen * mathb
      <10.95> mathb10 <12> <14.4> <17.28> <20.74> <24.88> mathb12
      }{}

\newcommand*{\Neptune}{{\text{\mathabxbfamily\char"48}}}

\newcommand*{\wasyfamily}{\fontencoding{U}\fontfamily{wasy}\selectfont}

\newcommand*{\mercury}{{\text{\wasyfamily\char39}}}

\makeatletter
\def\@fnsymbol#1{\ensuremath{
\ifcase#1\or
\mercury \or  
\Neptune \or 
\mathsection \or
\mathparagraph \or
\|\or **
\or
\dagger\dagger
\or \ddagger\ddagger
\else\@ctrerr\fi}}
\makeatother

\begin{document}

\title{``Hunt Takes Hare'': Theming Games Through Game-Word Vector Translation}


\author{Youn\`{e}s Rabii}
\affiliation{%
  \institution{Queen Mary University of London}
  \city{London}
  \country{UK}
}
\email{yrabii.eggs@gmail.com}

\authornote{References to this author must be made using the they/them singular neutral pronouns.}

\author{Michael Cook}
\affiliation{%
  \institution{King's College London}
  \city{London}
  \country{UK}
}
\email{mike@possibilityspace.org}

\authornote{References to this author must be made using he/him or they/them singular neutral pronouns.}

\renewcommand{\shortauthors}{Rabii and Cook}
\renewcommand{\shorttitle}{``Hunt Takes Hare'': Theming Games Through Game-Word Vector Translation}

\begin{abstract}
A game's theme is an important part of its design -- it conveys narrative information, rhetorical messages, helps the player intuit strategies, aids in tutorialisation and more. Thematic elements of games are notoriously difficult for AI systems to understand and manipulate, however, and often rely on large amounts of hand-written interpretations and knowledge. In this paper we present a technique which connects \textit{game embeddings}, a recent method for modelling game dynamics from log data, and \textit{word embeddings}, which models semantic information about language. We explain two different approaches for using game embeddings in this way, and show evidence that game embeddings enhance the linguistic translations of game concepts from one theme to another, opening up exciting new possibilities for reasoning about the thematic elements of games in the future.
\end{abstract}



\maketitle

\section{Introduction}
Almost a decade ago, Cook and Smith wrote that the `mechanics-first view on games is unnecessarily limiting' and called for AI research into game design to consider experiential, aesthetic and rhetorical aspects of games in more depth \cite{smithcook}. Despite new work along these lines, building automated systems that can understand, change or create the theming and contextual meaning for a videogame is an underexplored area of game AI research.

Understanding a game's theme requires the ability to relate game concepts to real-world ideas. This is useful for a wide variety of applications, including the contextualisation of generated content, game design assistance and guidance, and tutorialisation. However, an abstract game carries no indication of how its components should be themed, and a surface-level translation of an existing theme that does not take into account how a game is played will likely lose coherence and impact. 

Previous attempts to overcome this problem have tended to rely on pre-made databases of meaning, and user-defined graphs of concepts that relate to one another \cite{gemini}\cite{treanor}. We cover some of this work in section \S\ref{sec:related}, \textit{Related Work}. While this approach has many benefits, it is very time-consuming, does not scale or generalise, and cannot be extended to procedurally generated concepts for which no prior knowledge exists. 

In \cite{rabii}, Rabii et al present a technique which uses word embeddings to train a vector-space representation of game log data. They show that with no prior knowledge about the game other than the logs that it is possible to extract complex, high-level knowledge about the game's structure, dynamics and strategy. This approach circumvents some of the issues mentioned above, in that it produces knowledge about a game simply through observations of it being played. However, this knowledge is completely abstract -- it has no connection to the grounded `meaning' of the game.

In this paper we repurpose Rabii's work, and present our technique for using a combination of game log embeddings and word embeddings to translate a game's theme across a semantic space. We describe our system setup and how we relate game embeddings to the real-world linguistic knowledge base represented by word embeddings. We also describe our approaches to translating thematic elements across the word embedding space to yield new themes for games. We share some preliminary results here, and note the challenges we have encountered so far, and our preliminary steps to improve upon the work. This represents a starting point for a new application of word embeddings to help relate game dynamics to real-world ideas.

The remainder of this paper is organised as follows: in section \S\ref{sec:background} we cover how word and game embeddings work; in \S\ref{sec:related} we discuss related work in computational creativity and automated game design; in \S\ref{sec:methodology} we cover the methodology of our approach to thematic translation; in \S\ref{sec:results} we present and evaluate initial results from the system; and in \S\ref{sec:conc} we conclude our work and look to the future.

\section{Background}\label{sec:background}
\subsection{Word Embeddings}
Word embeddings are representations of sequential data (usually language) in the form of n-dimensional vectors, used in natural language processing for tasks such as topic modelling and semantic distance measurement. The process is trained on a dataset of token sequences, and the positions of tokens in the resulting vector space aims to mimic their distributions in the original data. For natural language, this means that words with a similar semantic role in the training data tend to appear close to each other along certain dimensions in the embedded space. Word embeddings are useful not only for measuring the distance between tokens, but also because mathematical operations can be performed on the vectors, such as addition, subtraction or averaging. This allows for calculations to be performed on words to compose or subtract meanings. Parrish gives the example of colours in \cite{parrish}: subtracting the vector for the word \textit{Blue} from the vector for the word \textit{Sky} and adding the result to the vector for the word \textit{Grass} yields a vector close to the word \textit{Green}. The best-known algorithm for creating word embeddings is Word2Vec, which has been widely studied both within computer science as well as beyond in digital arts and other creative fields \cite{mikolov_efficient_2013}. Word2Vec's simplicity and broad scope has made it a useful and accessible tool for artists and creative coders as well. 

\subsection{Game Embeddings}
In \cite{rabii} Rabii and Cook present an application of word embeddings to gameplay data. They take a set of formally-annotated chess match logs, and translate it to a domain-specific language that they designed for the purposes of training an embedding. They then apply the word2vec algorithm to this chess gameplay data, yielding a vector-space embedding for the game logs. In their paper they show that the resulting embedding reveals a range of interesting information about chess, from fundamental structural concepts through to subtle strategic insights understood by proficient players.

In their original proposal, the authors note several important traits of their system: the design of the domain-specific language, the size of the data, the source of data points (for example, expert players versus novices) and the exact phrasing of queries all have a significant impact on the types of conclusions that can be drawn from the data. Despite this, the work clearly shows that the embeddings capture important truths about the game simply by seeking semantic connections between logs of game events. This technique is potentially very scalable and widely-applicable as a result, since a game developer can easily gather event logs for their own game, and create a vector embedding from it. 

Both game and word embeddings represent relationships between concepts through a multi-dimensional vector space, learned from datasets. As stated above, we can operate within these vector spaces to transform and translate concepts, through processes like addition or subtraction. However, vector spaces can also be related to one another, particularly when we can provide points which overlap or connect semantically. We can draw connections between these spaces through the use of regression techniques \cite{rodriguez}, which allow us to use known connections between spaces to infer new ones. In this paper, we present our efforts to draw a connection between a game embedding space (trained on chess data, as per Rabii and Cook) and a word embedding space (trained on English language text) to show how linguistic and ludic concepts can be linked together, and then transformed to find new themes and meaning in word embedding space.

\section{Related Work}\label{sec:related}
\subsection{Metaphor and Analogy}
In this paper we seek to re-theme a game. We have a set of game elements, such as the pieces on a chessboard and the states and rules of the game, and we wish to rename these elements to align with a new theme. This has some similarity to the notion of metaphor or analogy in natural language processing research. A metaphor has two elements: a \textit{tenor}, the root concept that is being conveyed, and a \textit{vehicle}, a second concept that is used to represent the tenor in a new, metaphorical context. Metaphors usually connect concepts together through a specific connection point and then encourage the audience to extrapolate other, looser connections from there. 

Metaphor and analogy are a widely studied topic in broader AI and natural language processing research, especially in computational creativity where researchers such as Veale have extensively studied how computers can extract metaphors from language \cite{veale}. Veale's approach identifies common structures in natural language and mining those associations to create a large knowledge base. This is supported by existing linguistic knowledge graphs such as WordNet \cite{wordnet}. In \cite{gero}, Gero et al use the open-source knowledge graph GloVe to evaluate potential metaphors and grades different qualities of the metaphor based on their semantic similarity in the knowledge graph. Our approach also uses GloVe as a basis for its linguistic knowledge.

\subsection{Theming Games}

Theming or otherwise providing semantic meaning to game elements is primarily a topic covered by automated game design research, which Cook defines as `the science and engineering of AI systems that model, participate in or support the game design process' \cite{puck}. Such systems engage with the topic of game theming for a variety of reasons, including wanting to impose a theme on an existing game, or interpreting a game to understand possible themes it may be expressing. Treanor's work on micro-rhetorics for the Game-o-Matic system is one such example -- micro-rhetorics are user-defined components of meaning which can be combined with one another to compose higher-level rhetorical messages \cite{treanor}. In his work on the Game-o-Matic, Treanor combines micro-rhetorics with thematic graphs that users provide (for example, `cop arrests protester') to synthesise games which convey a theme.

GEMINI furthers work in this area, by providing bidirectional interpretation \cite{gemini}. This allows the GEMINI system to not only create games based on thematic structures, but also to intuit possible thematic structures by examining an unthemed game, which is closer to the task we approach in this paper. This is done using a pre-compiled database of structures and their rhetorical interpretations, which formulates the task of theming a game to a logic programming task using answer set solvers. Notably, GEMINI operates on rules, not behaviour -- `mechanics', rather than `dynamics', in the parlance of Hunicke et al \cite{mda} -- and its rhetorical databases are also defined in this way, describing the rhetorical meaning of game elements directly. Our approach considers the dynamic behaviour of game elements instead, by focusing on Rabii's game embeddings technique, and leverages large existing knowledge databases (in our case, word embedding models) that are not specific to any particular game.

\begin{table}[t]
    \centering
    \begin{tabular}{|p{0.45\linewidth}|p{0.45\linewidth}|}
        \hline \begin{spverbatim}Turn Black Rook C4 R7 C4 R0 Capture Knight\end{spverbatim}& \textit{Black moved a Rook from (4,7) to (4,0), capturing a Knight.} \\ \hline
        \begin{spverbatim}Turn Black Pawn C5 R1 C5 R0 Promote Queen Check\end{spverbatim}& \textit{Black moved a Pawn from (5,1) to (5,0), promoting it to a Queen. The game is in check.} \\ \hline
        \begin{spverbatim}Turn White King C4 R7 C2 R7 Castling Checkmate WhiteWin\end{spverbatim}& \textit{White moved its King from (4,7) to (2,7), with the single-use castling move. The game is in checkmate. White wins.} \\ \hline
    \end{tabular}
    \caption{Excerpts of chess games in our description language (left) and their natural language equivalent (right)}
    \label{tab:language_example}
\end{table}

\begin{table}[t]
\begin{tabular}{lll|l|l|}
\cline{1-2} \cline{4-5}
\multicolumn{1}{|l|}{\textbf{Token}} & \multicolumn{1}{l|}{\textbf{Words}} & \textbf{} & \textbf{Token} & \textbf{Words}  \\ \cline{1-2} \cline{4-5} 
\multicolumn{1}{|l|}{\texttt{White}}          & \multicolumn{1}{l|}{white}                &           & \texttt{Capture}        & capture               \\ \cline{1-2} \cline{4-5} 
\multicolumn{1}{|l|}{\texttt{Black}}          & \multicolumn{1}{l|}{black}                &           & \texttt{Castling}       & castling              \\ \cline{1-2} \cline{4-5} 
\multicolumn{1}{|l|}{\texttt{King}}           & \multicolumn{1}{l|}{king}                 &           & \texttt{Promote}        & promote transform     \\ \cline{1-2} \cline{4-5} 
\multicolumn{1}{|l|}{\texttt{Queen}}          & \multicolumn{1}{l|}{queen}                &           & \texttt{Checkmate}      & checkmate             \\ \cline{1-2} \cline{4-5} 
\multicolumn{1}{|l|}{\texttt{Bishop}}         & \multicolumn{1}{l|}{bishop}               &           & \texttt{Check}          & check, control, prevent \\ \cline{1-2} \cline{4-5} 
\multicolumn{1}{|l|}{\texttt{Rook}}           & \multicolumn{1}{l|}{rook}                 &           & \texttt{Stalemate}      & stalemate, deadlock    \\ \cline{1-2} \cline{4-5} 
\multicolumn{1}{|l|}{\texttt{Knight}}         & \multicolumn{1}{l|}{knight}               &           & \texttt{WinWhite}       & victory               \\ \cline{1-2} \cline{4-5} 
\multicolumn{1}{|l|}{\texttt{Pawn}}         & \multicolumn{1}{l|}{pawn}                 &           & \texttt{WinBlack}       & defeat                \\ \cline{1-2} \cline{4-5} 
                                     &                                           &           & \texttt{Draw}           & draw, tie, deadlock     \\ \cline{4-5} 
\end{tabular}
\caption{List of tokens and their associated words used to define Chess's narrative theme}
\label{tab:chess-theme}
\end{table}

\section{Methodology}\label{sec:methodology}

Our goal is to build a system that can associate words in English to Chess game concepts, by relating English word embeddings to game embeddings built from Chess play data. If $token$ is a game token, we note $G_{token}$ its associated game vector.
If $word$ is a word in our dataset's English vocabulary, we note $W_{word}$ its associated word vector.
If $w_1, w_2 \ldots ,w_n$ are words in our dataset's English vocabulary, we note $W_{w_1,w_2...,w_n}$  the average word vector $\sum_{i=1}^{n} \frac{w_i}{n}$. The relationship vector between two words $w_1$ and $w_2$ is denoted ${W_{w_1 \to w_2}} = W_{w_2} - W_{w_1}$. This is a vector which can be `added' to the vector $W_{w_1}$ to yield the vector representing $W_{w_2}$.

\subsection{Training Embeddings}
Our source for word vectors are the public domain GloVe pre-trained embeddings, trained on the Wikipedia 2014 and the Gigaword 5 corpus \cite{pennington2014glove}. Its vocabulary contains 400,000 words, each represented by a 50-dimensional vector.

To create our game embeddings, we reproduced Rabii and Cook's  \cite{rabii} data processing pipeline. We first downloaded a month of ranked matches data from the online Chess playing platform Lichess, and converted each logged game in the custom description language described by the authors. Samples of the converted data with their natural language equivalent are provided in Table \ref{tab:language_example}. 

The description language contains a total of 34 different tokens that represent different concepts used to describe a game of Chess. Tokens can represent players (e.g. \verb|White, Black|), Chess pieces (\verb|Queen, Pawn|), playing moves (\verb|Capture, Castling|), game states (\verb|Checkmate, WinBlack|) and even the eight rows and columns of the board (e.g. \verb|R0, R1, R7, C0, C1, C7|).

We then used our corpus of Chess data to train a Word2Vec model using the same settings as Rabii and Cook. In the end, each of our 34 Chess tokens has a 5-dimensional vector representation.

\subsection{Themes and Rethemings}

We define a \textit{theming} as a function that associates game tokens built from our Chess description language to word vectors in our dataset.

We consider Chess to have an already established theming, which we denote as the function $theme$. Our definition of Chess' current theme can be found in Table \ref{tab:chess-theme}. It does not contain every token in the original description language, only those that we could associate to a specific word vector. For example, since we associate the game token \verb|King| to the word "\textit{king}", we have:
$$theme(\verb|King|) = W_{king}$$
However, $theme$ cannot take as input the tokens that represent the rows and columns of the board (\verb|R0,...,C7|), as we couldn't map them to a specific word in English. Some tokens can be associated to a word, but that word might have other meanings than the one we want to use (e.g. \verb|Draw| does not refer to the act of drawing a picture, but a situation where no player wins). In that case, we associate the token to at most three words $w_1,..,w_3$, and use the average of their word vectors $W_{w_1,w_2,w_3}$. The set of tokens that are valid inputs for $theme$ is called $\mathcal{T}$.  

Our goal is to \textit{retheme} Chess, i.e. create other themings with alternative token-to-words associations.
We define $retheme$ as a function that, given a token and a relationship vector $start \to end$,  returns a word vector:
$$\underset{{start} \rightarrow {end}}{retheme}(token) = W_{word}$$
The relationship vector guides the semantic translation of the input token into the final output word. For example, if we built a function to shift the masculine words in Chess' theming to their feminine equivalents, we might have the following equations:

$$theme(\verb|King|) = W_{king}$$
$$ \underset{{masculine} \rightarrow {feminine}}{retheme}(\verb|King|) = W_{queen}$$

In the following sections, we describe three different models for building the $retheme$ function, one using only word vectors, and two using a combination of word and game vectors. In the following results section, we show and discuss results from all three approaches.


\subsection{Word Vectors Only}\label{sec:words-only}
Our first retheming model is built using the well-known Word2vec's analogy formula, adapted to the general case:
$$\underset{{start} \rightarrow {finish}}{retheme}(token) = theme(token) + W_{start \to finish}$$ 
For the above example of translating across the theme vector of masculine to feminine, we would therefore calculate the retheming as:
$$W_{queen} = W_{king}  + W_{masculine \to feminine}$$
This function only uses word vectors. We chose it as a baseline to compare with techniques that leverage the information stored in game vectors.

\begin{table*}[t]
\centering
\begin{tabular}{lllllll}
\toprule
Token & Queen & Rook & King & Bishop & Knight & Pawn \\
\toprule
\textbf{Expert valuation (Evans, 1958)} & 10.0 & 5.0 & 4.0 & 3.75 & 3.5 & 1.0 \\
Similarity to token "\texttt{Checkmate}" (game vectors) & 10.0 & 9.96 & 9.41 & 5.99 & 4.93 & 1.0 \\
Similarity to token "\textit{checkmate}" (word vectors) & 2.91 & 7.36 & 3.49 & 1.0 & 5.64 & 10.0 \\
Similarity to token "\textit{win}" (word vectors) & 9.61 & 1.0 & 10.0 & 4.42 & 9.84 & 5.58 \\
\bottomrule
\end{tabular}
\caption{All values are linearly normalised to be between 1 and 10.}
\label{tab:valuation}
\end{table*}

\subsection{Using Both Game and Word Vectors}\label{sec:game-and-word}

Game embeddings contain expert knowledge about the specific game they were extracted from, while word embeddings represent the meaning of words in the broad context of their dataset. For example, some expert players assign a numeric valuation to each chess piece as a way to quantify how important they are for winning a game. If we compute the cosine similarity between $G_{King}, G_{Queen}, ... G_{Pawn}$ and the vector $G_{Checkmate}$, we can extract an ordering of chess pieces which aligns with the ordering of importance as perceived by chess masters. When doing the same basic operation with word vectors instead, we found no sign of correlation with the expert valuation of chess pieces. Table \ref{tab:valuation} shows these comparisons. This suggests that game embedding data carries important information about the structure of Chess as a \textit{game} which the words alone do not. Therefore, we hypothesize that a retheming using both word and game vectors can provide us with more relevant results than ones purely based on word vectors. 

We want our model to be able to translate relationships between game tokens (such as the relative value ordering of chess pieces) into relationships between words (e.g. the medieval hierarchy of Chess piece names). We are interested in models of the form:
$$\underset{{start} \rightarrow {finish}}{retheme}(token) = f(G_{token}) + W_{start \to finish}$$
where $f$ is a linear function that associates a game vector to a word vector. The choice of linear functions is motivated by the fact that semantic relationships between words and game concepts are captured in relationship vectors $G_{1 \to 2}$ or $W_{1 \to 2}$. \cite{rabii} \cite{analogies_explained} A linear transformation $f: X \rightarrow \alpha X+\beta$ preserves relationship vectors up to a scaling factor $\alpha$.
$$f(G_{1 \to 2}) = f(G_2) - f(G_1) = \alpha W_2 - \alpha W_1 = \alpha  W_{1 \to 2}$$


\noindent Recall that $\mathcal{T}$ is the set of tokens which are valid inputs for the theming function. For each game token $T \in \mathcal{T}$, we train a linear regression model $f_T$ on a subset $\mathcal{T}_{N,T}$ of $N$ game tokens randomly chosen in $\mathcal{T}$ and their corresponding value given by $theme$. The purpose of $f_T$ is to convert the gameplay relationships embedded in the game vector $G_T$ to semantic relationships, encoded in a word vector. Unlike the method presented in Section \ref{sec:words-only}, we don't want the prediction to be influenced by the currently given value of $theme(G_T)$ so we make sure that the token $T$ is absent from $\mathcal{T}_{N,T}$.
$$\begin{array}{ccc} \mathcal{T}_{N,T} \subseteq \mathcal{T} & T \notin \mathcal{T}_{N,T} & |\mathcal{T}_{N,T}| = N\\ \end{array}$$

$${(\alpha_{T}, \beta_{T}) = \underset{(\alpha,\beta)}{\mathrm{Argmin}}  \sum_{t \in \mathcal{T}_{N,T}} \| \alpha G_{t}+\beta-{theme}(t)\|^{2}}$$

$$f_T(G_T) = \alpha_T G_{T} + \beta_T$$

\noindent Once the conversion from the game to word embeddings space is done, we add the relationship vector that guides the retheming:

$$\underset{{start} \rightarrow {finish}}{retheme}(T) = f_T(G_T) + W_{start \to finish}$$

\subsection{Choosing the guiding vector}

The $retheme$ function requires a relationship word vector of the form $W_{start \to finish} = W_{finish} - W_{start}$, that we call the guiding vector. It is best to think of it as representing a semantic transformation from the starting word $start$ to the target word $finish$. In our experiments, this transformation is applied to the narrative context of Chess, to shift it from its current theme to a different target theme. We experimented with two types of guiding vector. 
\subsubsection{Guiding based on a specific example}\label{par:specific-vector}

If we have a token in our Chess description language that we want to retheme to a specific word (eg. map the token \verb|King| to the word \textit{lion} instead of the word \textit{king}), we can provide the vector representing that exact transformation: $W_{king \to lion}$. All the other rethemed word vectors will be translated by that same vector. We can think of this translation as being anchored around the given example, with the other translations being secondary and hopefully following the same overall trajectory.

\subsubsection{Guiding based on a target semantic field}\label{par:semantic-field}

Instead of providing a specific retheming for a token ($king \to lion$), we can choose a broader semantic field, represented by a list of words $B = b_1, b_2, ...b_n$. Let $A = a_1, a_2, ..., a_m$ be the words used in the current theming of Chess in Table \ref{tab:chess-theme}. We can create a vector representing the shift from one lexical field to another by first computing the average vector of each list -- respectively $W_{B} = \frac{1}{n} \sum_{i=1}^{n} W_{b_i}$ and $W_{A} = \frac{1}{m} \sum_{i=1}^{m} W_{a_i}$. The guiding vector $W_{A \to B} = W_{B} - W_{A}$ represents the transformation of shifting from the starting theme to the one specified by the list of words given in input.

\begin{table*}
\begin{tabular}{llll}
\toprule
Tokens & 1. Word vectors only & 2. Word+Game (N=10) & 3. Word+Game (N=5) \\
\midrule
\texttt{Black} & blue black pink & foxes grizzly hyena & blue white green \\
\texttt{White} & blue black green & lion grizzly bear & black blue pink \\
\texttt{Bishop} & saltire borough primate & hyena heelers furry & grayish striped silvery \\
\texttt{King} & lion dragon elephant & swan queen princess & continentwide free-range bizonal \\
\texttt{Knight} & lion sable grizzly & striped underside blue & suffragan ripon archbishop \\
\texttt{Pawn} & pawn man-sized henhouse & lion sable hyena & black blue green \\
\texttt{Queen} & lion mermaid dragon & whelp bullseye passant & black blue green \\
\texttt{Rook} & rook dippin maned & bear small black & lion lady queen \\
\texttt{Check} & mosquito trap sniffing & lion boar dragon & lion swan dragon \\
\texttt{Checkmate} & whelp t-1000 airborn & snow white climate & preyed genovese villainous \\
\texttt{Draw} & fringe u.s.-controlled netting & staving staved dampener & pawn gajar dispater \\
\texttt{Stalemate} & staving u.s.-controlled continentwide & capering side-chain polyelectrolyte & pawn gumdrop bejeweled \\
\texttt{WinBlack} & underdogs wolves relegation & tied tie bottom & stormy impasse mbeki \\
\texttt{WinWhite} & victory 1-0 2-1 & underdogs relegation rout & decisive defeat rout \\
\texttt{Capture} & capturing trap capture & lion grey sable & sable lion bloods \\
\texttt{Castling} & bow-tie ant-like tzitzit & lion rainbow blue & black blue white \\
\texttt{Promote} & empowerment sustainable promotes & ringlets hydrophobic quartile & metahumans angoulême 54-kg \\
\hline
Average R² & N/A & 0.61 ± 0.06 & 1.0 ± 0.0 \\
\bottomrule
\end{tabular}
\caption{Rethemings of game tokens using guiding vector \textit{King} $\rightarrow$ \textit{Lion}}
\label{king_lion_True_146702}
\label{tab:results-1}
\end{table*}

\begin{table*}
\begin{tabular}{llll}
\toprule
Tokens & 1. Word vectors only & 2. Word+Game (N=10) & 3. Word+Game (N=5) \\
\midrule
\verb|Black| & wild black squirrel & wild boar ant & mormon latter-day darwin \\
\verb|White| & wild green white & wild domesticated ant & indle great-grandson ahrts \\
\verb|Bishop| & latter-day apostle bishop & wild foxes hunters & knight grandson nephew \\
\verb|King| & king famous ancestor & cat lion butterflies & spread areas weeks \\
\verb|Knight| & hunter badger wolf & boar thornberrys pheasant & capture capturing expedition \\
\verb|Pawn| & pawn boar stinkhorn & wild hare native & dragon knight adventures \\
\verb|Queen| & queen enchanted famous & wild hunting hunters & spearpoint ftsz n-channel \\
\verb|Rook| & rook groening voboril & hunt wild famous & capture capturing elude \\
\verb|Check| & wild breeding cattle & bee gees narnia & famous queen wild \\
\verb|Checkmate| & eriogonum brettanomyces aphid & deadlock stalemate impasse & cattle wild sheep \\
\verb|Draw| & wild ducks 2-2 & wild disastrous logging & black white gray \\
\verb|Stalemate| & spawning stalemate ftaa & bcg spiritism self-sustaining & darwin augustine lds \\
\verb|WinBlack| & wild defeat 4-0 & victory wild 4-0 & rook ajaw vålerengen \\
\verb|WinWhite| & victory wild 4-0 & wild defeat tigers & mccurry mcclellan fratto \\
\verb|Capture| & hunters wild hunt & ant wild asteraceae & mittelfranken r2000 k21 \\
\verb|Castling| & castling pseudotooth mouflon & wild ghost hunter & nelson clive muir \\
\verb|Promote| & breeding ecotourism transgenic & tackie amed jan.-may & stalemate deadlock impasse \\
\hline
Average R² & N/A & 0.6 ± 0.03 & 1.0 ± 0.0 \\
\bottomrule
\end{tabular}
\caption{Rethemings of game tokens using a target semantic field related to "wildlife"}
\label{tab:results-2}
\label{face_wild_True_12159}
\end{table*}

\section{Results}\label{sec:results}
The models described in Section  \S\ref{sec:methodology} were built with the goal of retheming the tokens in our Chess description language. We present our results under different modalities in Tables \ref{tab:results-1} and \ref{tab:results-2}. Each model is tasked with associating all the game tokens in Table \ref{tab:chess-theme} to a word embedding. We omit the exact values of each 50-dimensional vector here and instead give the top three closest words featured in our English dictionary, according to cosine distance.

We explore two types of retheming, each with the goal of shifting the original medieval warfare theming of Chess to a theme about wildlife. Table \ref{tab:results-1} contains the models output when prompted with the specific transformation $W_{king \to lion}$ (see \ref{par:specific-vector}). Table \ref{tab:results-2} contains results obtained when prompting the models with the target semantic field represented by the following words: \textit{lion, elephant, zebra, eating, becoming, extinction, control, win, loss}. Our explorations showed that ideally, this list should not only contain words that refer to animals, as words about the basic concepts of a strategy game (eg. \textit{win} and \textit{loss}) are important to define the semantic field of a Chess game.

Applying the procedure outlined in \ref{par:semantic-field} gave us a guiding vector $W_{A \to B}$ that is similar to the vector $W_{face \to wild}$, as the closest words to the average of each semantic field were "\textit{face}" for Chess, and "\textit{wild}" for the list of words related to wildlife.

For each modality, we compare the results of three models: a baseline retheming using only word vectors, followed by two rethemings using both word and game vectors combined, with the sampling parameter $N$ set to 5 and 10, respectively.

The maximum value for $N$ is the size of the original Chess theming in Table \ref{tab:chess-theme} minus one, i.e. 16. Our experiments using values of $N$ close to 16 yielded poor results, with little variation between each output. Choosing lower values of $N$ provides more diversity in the output word vectors, but increases the probability for each linear regression model to over-fit on the training set, which is evidenced by their average $R^2$ of 1 in the lower case where $N=5$. Over-fitting reduces the model's ability to generalize, leading to more words that are either too close the original theme (such as $knight$ or $queen$) or seem too far from the target theme (eg. $mormon$, $mittel-franken$, $mccurry$). We observed empirically that good trade-off seem to lie in the middle, with $N=10$. This trade-off may be different for other games and for differently-sized or -distributed token sets, which future work will investigate.

Comparing the baseline rethemings (Column 1) and the ones with $N=10$ (Column 2), we observe that models that don't use game vectors often provides words that are close, if not exactly the same as the starting theme ($king$ for \verb|King|, $pawn$ for \verb|Pawn|, $apostle$ for \verb|Bishop|) while the ones that do use them reuse few words from the original Chess theme.

For both modalities, the guiding vector's target words ("\textit{lion}" and "\textit{wild}" respectively) are often the closest word to the output of each model. This behavior was already observed in previous work on Word2Vec analogies \cite{levy2014linguistic} \cite{allen2019analogies}. As a result of this work, a common practice in analogy tasks is to simply discard target words from the results and use the second closest instead. We follow this principle in Step 2 of our retheming algorithm (\ref{sec:compare-pieces}).

\subsection{Comparing Retheming of Chess Pieces}\label{sec:compare-pieces}

We compare our $N=10$ model and the baseline model on the task of generating a set of chess pieces themed around wildlife. We use the results generated in Table \ref{tab:results-2}, columns 1 and 2. For each token related to a chess piece, we follow this algorithm:
\begin{enumerate}
    \item Compute the three closest words to the model's output.
    \item If the guiding vector's target word is in the list, discard it.
    \item Among the remaining words, pick the first noun of the list.
\end{enumerate}

The result of that process is presented in Table \ref{tab:retheme-pieces}. We note that the baseline model outputs words that are quite similar from the starting theme (Column 1), while the model using game vectors seems to correctly feature a "wildlife" theme (Column 2).

In Column 2, we observe that the medieval social hierarchy of Chess has been replaced by a food chain: weaker pieces are associated to prey animals while stronger pieces are associated to predators. Interpreting the mapping, we remark that the King is linked to a $cat$, an animal that is both weak and cherished, which bears some similarity to the King's role in chess.

The Rook is associated with \textit{hunt}, while the Bishop is associated with \textit{foxes}. The Queen's moveset allows it to move either like a Bishop or a Rook, and its associated word \textit{hunters} could be considered a composition of both the notion of a group hunt, represented by the Rook, and foxes which are solitary hunting animals, represented by the Bishop. Semantic blends such as this are necessarily fuzzy, but we find these examples inspirational and engaging.

We consider that our model's ability to provide a retheming whose semantic relationships seems to match, at least in an interpretive manner, with gameplay properties is a strong sign of its potential. The baseline model doesn't exhibit the same properties, suggesting that exploiting the information captured in game embeddings is valuable to explore the link between gameplay and narrative theming.

\begin{table}[t]
\begin{tabular}{|l|c|c|}
\hline
\textbf{Token} & \multicolumn{1}{l|}{\textbf{Word vectors only}} & \multicolumn{1}{l|}{\textbf{Themed (Word+Game)}} \\ \hline
\verb|King|           & king                                            & cat                                                \\ \hline
\verb|Queen|          & queen                                           & hunters                                            \\ \hline
\verb|Bishop|         & apostle                                         & foxes                                              \\ \hline
\verb|Rook|           & rook                                            & hunt                                               \\ \hline
\verb|Knight|         & hunter                                          & boar                                               \\ \hline
\verb|Pawn|           & pawn                                            & hare                                               \\ \hline
\end{tabular}
\caption{Outputs of our retheming models, when tasked with generating a Chess set that fits a "wildlife" theme}
\label{tab:retheme-pieces}
\end{table}

\section{Discussion and Conclusions}\label{sec:conc}
In this paper we presented preliminary results on the use of both word and game embeddings to model and translate thematic elements of games. We motivated our work by building on prior research into game embeddings for chess, and word embeddings for natural language, and showed how the two can be combined together in different ways to alter the theme of a game such as chess. We provided evidence that our technique works better than simply using word embeddings alone to retheme words related to a game, and we propose that this is because the game embeddings capture essential qualities of various game design components and this knowledge supports a richer translation.

We believe this approach has many promising and interesting features. For one, it captures the \textit{dynamics} of a game, rather than its static elements: game embeddings reveal, for example, the relative strength of chess pieces, even though this knowledge is not described anywhere in the game's rules. Most automated game design research focuses on static mechanical elements of games, and struggles to bridge to dynamic or emergent properties. This could be an exciting new way to approach the understanding, extraction and use of dynamic game elements.

Embeddings are also scalable and comparatively easy to both create and understand. Adding logging to a game to record important events and create datasets, such as the one we used from chess matches, is a simple exercise no different to the debugging or playtesting that game developers already do. We have already worked with one independent game developer to add logging of this type to their game, and the process was clean and straightforward. Word2Vec arithmetic is a widely-used and taught technique among digital art communities and creative coders. This can be furthered through better tools build specifically to make training and exploring embedding spaces more amenable. This makes it a promising technique for widespread adoption in the games industry.

We are planning to experiment further with this technique on more traditional digital games, to assess how it performs on different kinds of game scenario and different kinds of game logs. At the time of writing, we are working with an independent game developer to gather data from one of their games, from both expert and novice players. We also hope to experiment with more ambitious uses for the technique as well, such as the ability to create names for game elements that have been procedurally generated at runtime. We believe there are many exciting applications for this basic technique waiting to be discovered.

\begin{acks}
    We would like to thank the reviewers for their helpful feedback and reflections on the paper. While we weren't able to include all of their suggestions, this is an ongoing work and we will continue to develop the ideas and approaches here in future. This work was supported by the IGGI CDT, and the Royal Academy of Engineering.
\end{acks}

\bibliographystyle{ACM-Reference-Format}
\bibliography{biblio}

\end{document}